\documentclass[letterpaper, 10 pt, conference]{ieeeconf}  

\IEEEoverridecommandlockouts                              

\usepackage{blindtext}
\usepackage{epsfig} %
\usepackage{sidecap}
\usepackage[usenames, dvipsnames]{xcolor}
\usepackage{floatrow}
\usepackage{mathptmx} 
\usepackage{times} 
\usepackage{amsmath} 
\usepackage{amssymb}  
\usepackage{mathrsfs}
\usepackage{tikz}
\usepackage{soul}
\usepackage{bm}
\usepackage{cite}
\definecolor{dg}{RGB}{219, 235, 220}
\usetikzlibrary{positioning}
\tikzset{main node/.style={circle,fill=blue!20,draw,minimum size=1cm,inner sep=0pt},
            }
\tikzset{head node/.style={circle,fill=dg,draw,minimum size=1cm,inner sep=0pt},
            }
\usepackage[utf8]{inputenc}
\DeclareUnicodeCharacter{2212}{$-$}

\usepackage{subcaption}

\usepackage{algorithm, algorithmic}

\usepackage{makecell}
\usepackage[hyperindex,breaklinks]{hyperref}

\usepackage{bbm}
\usepackage{booktabs}
\usepackage{multirow}
\usepackage{siunitx}
\usepackage{subfloat}
\usepackage[amsmath,thmmarks,hyperref]{ntheorem}[1.33]
\usepackage{tikz}
\usetikzlibrary{shapes, arrows, automata, plotmarks}
\usepackage[symbol]{footmisc}

\renewcommand{\thefootnote}{\fnsymbol{footnote}}

\input{mysymbol.sty}
\DeclareMathAlphabet{\mathcal}{OMS}{cmsy}{m}{n}

\title{\LARGE \bf
Learning Connectivity for Data Distribution in Robot Teams
}

\author{Ekaterina Tolstaya$^{1 \star}$, Landon Butler$^{1 \star}$, Daniel Mox$^{2}$, James Paulos$^2$, Vijay Kumar$^{1,2}$, Alejandro Ribeiro$^{1}$ 
\thanks{$^{\star}$ Indicates equal contribution.}%
\thanks{$^{1}$Dept. of Electrical and Systems Eng.,
University of Pennsylvania, USA
        {\tt\small \{eig,landonb3\}@seas.upenn.edu}}%
\thanks{$^{2}$Dept. of Mechanical Eng. and Applied Mechanics, University of Pennsylvania, USA}%
\thanks{Supported by ARL Grant DCIST CRA W911NF-17-2-0181, NSF Grant CNS-1521617,
ARO Grant W911NF-13-1-0350, ONR Grants N00014-20-1-2822 and
ONR grant N00014-20-S-B001, and
Qualcomm Research. We gratefully acknowledge the support of NVIDIA Corporation with the donation of the DGX-1 used for this research. The first and third authors acknowledge support from the National Science Foundation Graduate Research Fellowship Grant No. DGE-1845298.}
}

\begin{document}

\maketitle
\thispagestyle{empty}
\pagestyle{empty}

\begin{abstract}
Many algorithms for control of multi-robot teams operate under the assumption that low-latency, global state information necessary to coordinate agent actions can readily be disseminated among the team. However, in harsh environments with no existing communication infrastructure, robots must form ad-hoc networks, forcing the team to operate in a distributed fashion. To overcome this challenge, we propose a task-agnostic, decentralized, low-latency method for data distribution in ad-hoc networks using Graph Neural Networks (GNN). Our approach enables multi-agent algorithms based on global state information to function by ensuring it is available at each robot. To do this, agents glean information about the topology of the network from packet transmissions and feed it to a GNN running locally which instructs the agent when and where to transmit the latest state information. We train the distributed GNN communication policies via reinforcement learning using the average Age of Information as the reward function and show that it improves training stability compared to task-specific reward functions. Our approach performs favorably compared to industry-standard methods for data distribution such as random flooding and round robin. We also show that the trained policies generalize to larger teams of both static and mobile agents.

\end{abstract}

\section{Introduction} \label{sec:intro}

Large scale swarms of robots have demonstrated utility in solving many real world problems including rapid environmental mapping \cite{thrun2000real, thrun2005multi}, target tracking \cite{schlotfeldt2018anytime}, search after natural disasters \cite{baxter2007multi, jennings1997cooperative} and exploration \cite{tolstaya2020multirobot, amigoni2017multirobot}. In many of these scenarios, robot teams must operate in harsh environments without existing communication infrastructure, requiring the formation of ad-hoc networks to exchange information. Furthermore, agent actions may take them out of direct communication range with a subset of the team so that packets must be relayed through intermediate nodes to reach their intended destination. Ad-hoc robot networks are implicitly decentralized. In spite of this, many algorithms for control of multi-robot teams operate under the assumption that low-latency, global state information necessary to coordinate agent actions can readily be disseminated among the team. While detail about how this data distribution task is accomplished is often scarce, its success is vital to the overall performance of the team. In this work, we address this challenge by providing a task-agnostic, decentralized, low-latency method for data distribution in an ad-hoc network using Graph Neural Networks. Our system enables existing centralized algorithms to be deployed in robot teams operating in harsh, real world environments.

The problem of data distribution in a team of robots bears a strong resemblance to the problem of packet routing in a mobile ad-hoc network (MANET). In our case, up-to-date global state information such as velocity, pose, or map data must be maintained at each robot so that they can choose appropriate control actions. Likewise, ad-hoc routing protocols in MANETs require each node maintain some knowledge of the state of the network so that they can choose the next hop a packet should take in its journey from source to destination. For proactive protocols, this is often accomplished by periodically flooding the network with link state information so that each node maintains a current understanding of the network topology \cite{williams2002comparison}. This process can be inefficient and strain the network with contention and packet collisions and as a result, many different flooding schemes have been developed to mitigate the so-called broadcast storm problem \cite{tseng2002broadcast}. As is made clear in Section \ref{sec:problem_statement}, our method draws inspiration from these flooding schemes and employs them as points of comparison during evaluation.

In this work, we assume the network is used exclusively for the data distribution task in question. As such, it makes sense to skip the regular link layer overhead necessary to operate an ad-hoc routing protocol and directly consider a data distribution protocol handling the application layer traffic of the robotic system. In essence, our data distribution protocol is itself a routing protocol which circulates the state information needed by the robots in the place of network state information needed for packet routing. While conventional approaches use some form of flooding to accomplish this task, most are based on heuristics designed to minimize the age of information and network overhead \cite{tseng2002broadcast}. 
There exists no clear optimal approach. Thus, we believe this problem is well suited to a data driven approach based on Graph Neural Networks paired with reinforcement learning.

Graph Neural Networks (GNNs) have shown great promise for learning from information described by a graph \cite{wu2020comprehensive, battaglia2018relational}. Recent works have used GNNs to generate heuristic solutions to a variety of multi-robot problems, such as path planning \cite{chen2019learning, battaglia2018relational, joshi2019efficient},  exploration \cite{chen2019self}, and perimeter defense \cite{paulos2019decentralization}. In this work, a GNN is the natural choice for parameterizing the communication policy which prescribes actions based on the information exchanged over the network graph. Furthermore, GNNs offer attractive properties such as permutation invariance, making the learned solution robust to changes in graph topology between training and testing.

In order to effectively distribute information, our model must learn when and with whom to communicate. Many prior approaches use multi-agent reinforcement learning to address communication for particular cooperative control tasks, \cite{DBLP:journals/corr/SukhbaatarSF16, NIPS2016_c7635bfd,DBLP:journals/corr/abs-1812-09755, tolstaya2019learning}. These approaches generally do not incorporate bandwidth limitations or physical channel models, often assuming that all agents are reachable through broadcast communications, which is sometimes range-limited.  Other approaches learn communication graphs to keep communication sparse \cite{inala2020neurosymbolic, liu2020when2com}. However, such approaches begin creating these graphs by first assuming that the relative positions of all other agents can be noisily observed \cite{inala2020neurosymbolic} or are transmitted through a series of global broadcasts \cite{liu2020when2com}. In this paper, we consider realistic communication conditions and thus assume no prior knowledge of other agents except for their identities and overall team size.
 
The primary contribution of this work is the development of a learned, cross-layer data distribution protocol that can be applied to any multi-agent task that relies on sharing time-sensitive local observations across a wireless network. Using reinforcement learning, this protocol can be optimized for use with static or mobile teams. 

The remainder of the paper is organized as follows: Section \ref{sec:problem_statement} describes the wireless communication model, local data structures and the AoI minimization problem. Section \ref{sec:methods} details the GNN architecture, the design of the transmission-response protocol, and our usage of reinforcement learning. Section \ref{sec:simulations} describes the agent dynamics, the mission specification for the applications of our approach, and we compare  performance of the GNN to several baselines in Section \ref{sec:results}. An open-source implementation of our work can be found 
\href{https://github.com/landonbutler/Learning-Connectivity}{in this \textbf{repository}}.

\section{Problem Formulation}\label{sec:problem_statement}

\subsection{Wireless Communication Model}\label{sec:communication_model}

In order to reason about transmitting information in a network, it is necessary to establish a model for communication. In this work, we consider ideal wireless conditions between agents with line of sight channels and successful packet reception subject to interference from other transmitting nodes in proximity. Transmitted signals propagating through the air experience free-space path loss $\mathcal{E}_t^{i,j}$ given by:
\begin{equation}\label{eqn:fspl}
\mathcal{E}_t^{i,j} = 20\log_{10}(d_t^{i,j})+20\log_{10} (\nu) - 147.55
\end{equation}
where $d_t^{i,j}$ is the distance between transmitting agent $i$ and receiving agent $j$, and $\nu$ is the system center frequency (in Hz) \cite{goldsmith2005wireless}. This approach is consistent with wireless studies involving transmissions between aerial agents \cite{8254715,7037248}.
From path-loss, we can define the channel gain as:
\begin{equation}\label{eqn:channel-gain}
g_t^{i,j} = 10^{-\frac{\mathcal{E}_t^{i,j}}{10}}
\end{equation}
and the signal-to-interference-plus-noise ratio (SINR) as:
\begin{equation}\label{eqn:sinr}
\Gamma_t^{i,j} = \frac{\rho_t^i\cdot g_t^{i,j} }{\sigma + \sum\limits_{k \in \mathcal{A}\setminus i} \rho_t^k\cdot g_t^{k,j}}
\end{equation}
where $\sigma$ is additive white Gaussian noise representing ambient noise power at the receiver and $\rho_t^i$ is the transmit power of agent $i$. In order for a packet transmitted from agent $i$ to be successfully received at agent $j$, $\Gamma_t^{i,j}$ must exceed a SINR threshold $\hat{\Gamma}$ that takes into account the QoS of the link \cite{shorey2006mobile}. Defining $R_t^i$ as the set of agents that receive a transmission from agent $i$ at time $t$, we can express the probability of successful packet reception at $j$ as:
\begin{equation}\label{eqn:sinr-thresh}
p(j \in R_t^i) =
 \begin{cases} 
1 & \Gamma_t^{i,j} \geq \hat{\Gamma} \\
0 & \Gamma_t^{i,j} < \hat{\Gamma} \\
\end{cases}
\end{equation}
Note that in ad-hoc networks, it is common for all nodes to contend for the same medium and thus be able to decode any packet transmission for which they are in range, regardless of if they are the intended recipient. In fact, this is a foundational aspect of many flooding algorithms \cite{williams2002comparison}. We also allow agents to eavesdrop on each others transmissions.

\subsection{Local Data Structures}

Critical to any networking protocol is the metadata it caches for decision making.
In our approach, this information forms the feature vectors consumed by our communication policy. And, since this is a data distribution task, there is an additional payload: given a set of robots $\mathcal{A}$ with clocks synchronized at time $t$, each robot maintains a data structure that contains agent $i$'s knowledge about agent $j$ indicated by $M^{i,j}_t$. This information is required by the robot to make decisions to accomplish some task. Along with the last known state, agent $i$ also stores a timestamp for this observation of the state of agent $j$, $T_t^{i,j}$. Agents can observe their own local states at the current time step, $\bbx_t^i$, and update their own memory accordingly:
\begin{equation}
M_t^{i,i} := \bbx_t^i, \quad T_t^{i,i} := t  \quad\quad \forall  i \in \mathcal{A}, \hspace{1mm} t \geq 0  
\end{equation}
Agents may make a decision to transmit their local data to others. The set $S_t^i$ represents the intended recipients of a transmission by agent $i$ at time $t$. In this work, we constrain the set of intended recipients to be at most one other agent. As outlined in the previous section, agents in proximity that transmit simultaneously run the risk of interfering with each other causing communication failures. Thus, we define $f$ as the stochastic communication model that determines the probability of receipt $p(j \in R_t^i)$ of agent $i$'s transmission by agent $j$ at time $t$:
%
\begin{equation}
\left\lbrace p\left(j \in R_t^i\right) \right\rbrace_{i \in \mathcal{A}} = f\left(\left\lbrace S_{t}^i , \bbx_t^i \right\rbrace_{i \in \mathcal{A}} \right), 
\end{equation}
where $f$ is the communication model described in Eq.  \ref{eqn:sinr-thresh}. 
We also record the timestamp at which agent $i$ last attempted communication with agent $j$, $L_t^{i,j}$:
\begin{align}
 j \in S_{t}^i  & \Longrightarrow L_t^{i,j} = t.
\end{align}
%
%

As data is transmitted by agents in the team, we can track how it propagates through the network as shown in the example in Fig. \ref{fig:buffer_diagram}.  
We define the parent reference notation $P_t^{i,k} = j$ to indicate that agent $k$ first passed its information to agent $j$ on its way to agent $i$.
If agent $i$ directly receives agent $k$'s information from $k$, then $P_t^{i,k} = i$. When agent $i$ receives a transmission from agent $j$ containing agent $k$'s state, agent $i$ can record the link along which agent $j$ had obtained that information, $P_t^{i,k} = P_t^{j,k}$. This record of transmission relationships  describes how data flowed between agents to construct agent $i$'s knowledge of the team's state at time $t$. If agent $i$ has received agent $j$'s transmission, then it will update its data structure if any of the received information is newer than its current data, $ \forall k \in \mathcal{A} $:
\begin{align}
 T_t^{i,k} < T_t^{j,k}, j \in R_{t}^i  & \Longrightarrow \\ \nonumber  
 &(T_t^{i,k} = T_t^{j,k})  \wedge  (M_t^{i,k} = M_t^{j,k})   \wedge (P_t^{i,k} = P_t^{j,k}) 
\end{align}

\subsection{Local Policy to Minimize Age of Information}

To effectively disseminate data in the network, a communication policy is installed at every node that operates entirely off of the information outlined in the previous section. More formally, a policy $\pi$ selects a recipient for its transmission using the set of information available to agent $i$ at time $t$: $\left\lbrace T_t^{i,j}, M^{i,j}_t, P^{i,j}_t, L^{i,j}_t\right\rbrace_{j \in \mathcal{A}}$. The same local stochastic communication policy $\pi$ is to be used by all agents:
\begin{equation}
p(k \in S_{t}^i) = \pi\left(\left\lbrace T_t^{i,j}, M^{i,j}_t, P^{i,j}_t, L^{i,j}_t\right\rbrace_{j \in \mathcal{A}}  \right) \quad \forall i, k \in \mathcal{A}
\end{equation}
Given the dynamic nature of robot teams, it is vital that each node maintain up-to-date information. Thus, the stochastic communication policy $\pi$ needs to minimize the average age of information (AoI) across all agents, where $t - T_t^{i,j}$ is the age of information for $i$'s knowledge of $j$'s state. To obtain the performance criteria for the whole system, we average over all agents, $\mathcal{A}$, the mission duration, $\mathcal{T}$, and team dynamics, $\mathcal{X}$:
\begin{equation} \label{eq:minimize_aoi}
\min_\pi \;\;  \mathbb{E}_{t \in \mathcal{T}, \;\; i \in \mathcal{A}, \;\; \bbx_t^i \in \mathcal{X}} \left\lbrack t - T_t^{i,j}  \right\rbrack
\end{equation}
\begin{figure}[t]
\centering

\resizebox{\textwidth}{!}{%
  \begin{tikzpicture}
    
    \draw [line width=0.5mm](-3.45,-3.0) -- (1.7,-3);
    \draw [line width=0.5mm](6.3,-3.0) -- (11,-3.0);
    \draw [line width=0.5mm](1.7,-2.3) -- (1.7,-3.7);
    \draw [line width=0.5mm](6.3,-2.3) -- (6.3,-3.7);
    \draw [line width=0.5mm](6.3,-3.0) -- (11,-3.0);
    \draw [line width=0.5mm](4,-3.7) -- (4,-9.0);
    \draw [line width=0.5mm](1.7,-2.3) -- (6.3,-2.3);
    \draw [line width=0.5mm](1.7,-3.7) -- (6.3,-3.7);
    \draw [line width=0.5mm](4,2.6) -- (4,-2.3);
    \node[inner sep=0pt] (russell) at (0,-13)
    {
    \Large
    \setlength{\tabcolsep}{0.5em} 
{\renewcommand{\arraystretch}{1.25}
\begin{tabular}{|c|c|c|c|c|}
\hline
\multicolumn{5}{|c|}{\textbf{Agent 0's Local Data Structure *}}         \\ \hline
\textbf{ID} & \textbf{TS} & \textbf{State} & \textbf{Parent} & \textbf{LC}\\ \hline
\textbf{0}  & 4          &    $M_4^{0,0}$          &       &  \\ \hline
\textbf{1}  & 2          & $M_4^{0,1}$            &     0  & 3 \\ \hline
\textbf{2}  & 2           & $M_4^{0,2}$            &    3   &  \\ \hline
\textbf{3}  & 4           & $M_4^{0,3}$            &   0    &   \\ \hline
\textbf{4}  & 3           & $M_4^{0,4}$            &    3   & 1 \\ \hline
\end{tabular}}};
    \node at (4,-16.2) {\LARGE\textbf{Tree Representation of Agent 0's Local Data Structure at $\mathbf{t=4}$}};
        \node at (0,-2.55) {\LARGE\textbf{$\mathbf{t=1}$}};
        \node at (8,-2.55) {\LARGE\textbf{$\mathbf{t=2}$}};
        \node at (0,-8.55){\LARGE\textbf{$\mathbf{t=3}$}};
                \node at (8,-8.55){\LARGE\textbf{$\mathbf{t=4}$}};
    \node at (4,-2.7) {\LARGE\textbf{Communication}};
    \node at (4,-3.3) {\LARGE\textbf{Rounds}};
    \node[head node] (01) at (0,2) {\Large0};
    \node[main node] (41) at (-1.902,.618) {\Large4};
    \node[main node] (11) at (1.902,.618) {\Large1};
    \node[main node] (31) at (-1.174,-1.618) {\Large3};
    \node[main node] (21) at (1.174,-1.618) {\Large2};
    
    \node[head node] (02) at (8,2) {\Large0};
    \node[main node] (42) at (6.098,.618) {\Large4};
    \node[main node] (12) at (9.902,.618) {\Large1};
    \node[main node] (32) at (6.826,-1.618) {\Large3};
    \node[main node] (22) at (9.174,-1.618) {\Large2};
    
    \node[head node] (03) at (0,-4) {\Large0};
    \node[main node] (43) at (-1.902,-5.382) {\Large4};
    \node[main node] (13) at (1.902,-5.382) {\Large1};
    \node[main node] (33) at (-1.174,-7.618) {\Large3};
    \node[main node] (23) at (1.174,-7.618) {\Large2};
    
    \node[head node] (04) at (8,-4) {\Large0};
    \node[main node] (44) at (6.098,-5.382) {\Large4};
    \node[main node] (14) at (9.902,-5.382) {\Large1};
    \node[main node] (34) at (6.826,-7.618) {\Large3};
    \node[main node] (24) at (9.174,-7.618) {\Large2};
    
    \node[head node] (05) at (8,-11) {\Large0};
    \node[main node] (45) at (6.098,-12.382) {\Large4};
    \node[main node] (15) at (9.902,-12.382) {\Large1};
    \node[main node] (35) at (6.826,-14.618) {\Large3};
    \node[main node] (25) at (9.174,-14.618) {\Large2};

    \path[draw,thick, line width=.9mm, ->]
    (01) edge node {} (41)
    (21) edge node {} (11)
    (12) edge node {} (02)
    (22) edge node {} (32)
    (03) edge node {} (13)
    (43) edge node {} (33)
    (34) edge node {} (04);
    
    \path[draw,thick, line width=1mm,loosely dotted,->]
    (45) edge node {} (35)
    (25) edge node {} (35)
    (15) edge node {} (05)
    (35) edge node {} (05);
\end{tikzpicture}
}
\caption{After four rounds of communication, agent 0 receives data from all other agents. Because parent references are also transmitted, agent 0 is able to use its data structure to reconstruct the spanning tree consisting of the most recent information of other agents in the network.
\vspace{-0.5cm}
}\label{fig:buffer_diagram}
\end{figure}

\hspace{-1mm}The objective of the data distribution problem is to minimize the average age of information across the team. Data distribution is a cooperative task, so we can provide the team with a single reward signal, rather than assigning rewards per agent. Furthermore, the training procedure is centralized, but the policy relies on only local information, so it can be deployed in a distributed system. 

We assume that the number and identities of robots in the team, $\mathcal{A}$, are known prior to mission execution. We also assume that robots share a common reference frame and a synchronized clock. Finally, we assume that communications are not bandwidth-limited relative to the size of the transmitted data, $\lbrace T_t^{i,j}, M^{i,j}_t, P^{i,j}_t\rbrace_{i,j \in \mathcal{A}}$. 
The memory required for each agent to store its knowledge of the system state is linear in the number of agents, $\mathcal{O}(|\mathcal{A}|)$, but system-wide, the memory and communication requirements are $\mathcal{O}(|\mathcal{A}|^2)$.

\let\thefootnote\relax\footnote{\hspace{-.9em}* ID : Agent Identity ($i$)\\
        \indent\indent\hspace{-.2em} TS : Last Observed Timestep ($T_t^{i.j}$)\\
        \indent\indent\hspace{.25em}State : Agent State Information ($M_t^{i.j}$)\\
        \indent\indent\hspace{.25em}Parent : Data Flow Parent ($P_t^{i.j}$)\\
        \indent\indent\hspace{.25em}LC : Last Attempted Communication ($L_t^{i.j}$)}\hspace{-.55em}

\section{Methods}\label{sec:methods}


\subsection{Aggregation Graph Neural Networks}

Our goal is to develop a policy for making communication decisions that each agent can evaluate using its local data structure. Recall that at each point in time $t$, every agent $i$ has cached the set $\left\lbrace T_t^{i,j}, M^{i,j}_t, P^{i,j}_t, L^{i,j}_t \right\rbrace_{j \in \mathcal{A}}$. By taking a graph perspective of the data, we can structure it in a way useful for learning. To begin, we form the set of node features as: $V^i_t = \left\lbrace T_t^{i,j}, M_t^{i,j},  L_t^{i,j}\right\rbrace_{j \in \mathcal{A}}$. The parent references $P_t^{i,j}$ capture adjacency relationships in the network, describing a tree structure from the perspective of each agent $i$. We can define a directed edge in this tree by the ordered tuple $(r_l, s_l)$, where $r_l$ and $s_l$ are the receiver and sender node, respectively, for the directed edge of index $l$. Then, the set of all directed edges in the graph of agent $i$ at time $t$ is $E^i_t = \left\lbrace\left(P_t^{i,k}, k\right)\right\rbrace_{k \in \mathcal{A} \setminus i}$. For the applications presented in this paper, input edge attributes are used, but at the intermediate layers, node features are aggregated using edge relationships. 
Putting it all together,
we denote the graph that represents agent $i$'s knowledge of the team at time $t$ as $\mathcal{G}_t^i = \left\lbrace E_t^i,V_t^i\right\rbrace$. Next, we outline our GNN architecture.

GNNs are an increasingly popular tool for exploiting the known structure of any relational system \cite{battaglia2018relational}.  In graph convolutional networks, the graph convolution operation is defined using learnable coefficients that multiply the graph signal by powers of the adjacency matrix \cite{kipf17-classifgcnn, gama2018convolutional}. We extend this architecture by incorporating non-linear graph convolution operations.

The building block of a GNN is the Graph Network Block. 
Given a graph signal, $\mathcal{G} =\left\lbrace  \{ {\bbe}_{n} \}, \{{\bbv}_l\} \right\rbrace$,  one application of the GN block transforms these features into $\mathcal{G}' = \left\lbrace \left\lbrace {\bbe}_{n}' \right\rbrace, \left\lbrace {\bbv}_l'\right\rbrace\right\rbrace$:
\begin{align} \label{eq:gn_block}
\bbe_n' = \phi^e(\bbe_n, \bbv_{r_n}, \bbv_{s_n}) , \;\; %
\bar{\bbe}_l' = \alpha^{e \rightarrow v}(E_l') , \;\;
\bbv_l' = \phi^v(\bar{\bbe}_l', \bbv_l).
\end{align}
$GN(\cdot)$ is a function of the graph signal $\mathcal{G}$, described by the application of $\phi^e$, $ \alpha^{e \rightarrow v}$ and $\phi^v$ in that order to produce the transformed graph signal $\mathcal{G}'$, with the same connectivity but new features on the edges and nodes. 

The aggregation operation $\alpha^{e \rightarrow v}$ takes the set of transformed incident edge features $E'_l=\left\lbrace \bbe'_n \right\rbrace_{r_n = l}$ at node $l$ and generates the fixed-size latent vector $\bar{\bbe}_l'$. Aggregations must satisfy a permutation invariance property since there is no fundamental ordering of edges in a graph. Also, this function must be able to handle graphs of varying degree, so the mean aggregation is particularly suitable to normalizing the output by the number of input edges \cite{tolstaya2019learning}:
\begin{equation} \label{eq:aggregation}
\alpha^{e \rightarrow v}(E_l') := \frac{1}{  |E'_l|} \sum_{\bbe_n' \in E'_l} \bbe_n'.
\end{equation}
%
The non-linear variant of the Aggregation GNN  uses learnable non-linear functions to update node and edge features:
\begin{align} \label{eq:nonlinear_gnn}
\phi_N^e(\bbe_n, \bbv_{r_n}, \bbv_{s_n}) & :=  \text{NN}_e( \lbrack \bbe_n, \bbv_{r_n}, \bbv_{s_n}\rbrack ), \\  \nonumber
\phi_N^v(\bar{\bbe}_l', \bbv_l)  & := \text{NN}_v( \lbrack \bar{\bbe}_l', \bbv_l \rbrack), 
\end{align}
where $\text{NN}_e$ and $\text{NN}_v$ are multi-layer perceptrons (MLPs). While a Graph Network Block can be used to compose a variety of architectures, for this work, we develop a variant of the Aggregation GNN in which the output of every GN stage is concatenated, and finally, processed by a linear output transform $f_{\text{out}}$ \cite{tolstaya2020multirobot, gama19aggGNN}.  
\begin{align} \label{eq:enc_proc_dec}
\mathcal{G}' = f_{\text{out}} \Big( \big\lbrack f_{\text{dec}} (f_{\text{enc}}(\mathcal{G})),  f_{\text{dec}} (GN(f_{\text{enc}}(\mathcal{G}))), \;\;\; \\ \nonumber  
\quad  f_{\text{dec}} (GN(GN(f_{\text{enc}}(\mathcal{G})))),  \;\; ...  \;\; \big\rbrack \Big)
\end{align}
The addition of the encoder $f_{\text{enc}}$ and decoder $f_{\text{dec}}$ layers was inspired by the Encode-Process-Decode architecture presented in \cite{battaglia2018relational}. The number of $GN$ operations is a hyperparameter and determines the \textit{receptive field} of the architecture, or how far information can travel along edges in the graph. A GNN with a receptive field of zero can be compared to the Deep Set architecture that neglects any relational data \cite{NIPS2017_f22e4747}. 
 

\subsection{Transmission-Response Protocol}

The GNN outlined in the previous section is responsible for deciding who a node should attempt to communicate with, if at all, at a given timestep. In this section we detail the communication protocol that executes these plans. The protocol is divided into two main steps: a transmission phase, following the output of the GNN as one might expect, and a response phase, where certain recipients of the transmission are able to respond. The addition of this response phase provides the transmitter with the ability to seek out information by targeting certain agents to exchange information with, a behavior that becomes valuable in the following section on reinforcement learning.

In the response phase, those agents that receive the packet for which they are the intended recipients, $ \forall j \text{ s.t. } (j \in R_{t}^i) \land (j \in S_{t}^i) $, are able to respond by transmitting a message back to the 
set of original transmitter(s), indicated by $\bar{S}_{t}^j$. Note that at all times an agent can only receive at most one transmission at a time. This is a fundamental limitation in ad-hoc networks with a shared medium. Furthermore, two transmissions targeting the same destination will likely result in a packet collision with no data successfully decoded at the receiver. In the response phase, we assume that if an agent is able to respond, it will attempt to do so. Recall also from Section \ref{sec:communication_model} that agents other than those who are the designated recipients can also decode transmitted messages and update their data structures. This set of agents, called \textit{eavesdroppers}, are described by $\forall j \text{ s.t. } (j \in R_{t}^i) \land ( j \notin S_{t}^i )$.


We summarize the design of our protocol in Alg. \ref{alg:protocol}. A communication time window $t$ consists of both a transmission and a response phase. At the beginning of a window, agents first update their own state using local observations. Then, they pass their updated data structure into the local communication policy to output a set of intended recipients for their transmission. If this set consists only of the agent itself, the agent chooses not to communicate during the given transmission phase. Agents then transmit to their intended recipients. For communications that are received above the SINR threshold, agents will compare the contents of the communication to their own local data structure, updating it with any new information received. Once data structure updates from the transmission phase conclude, the response phase commences. An agent will then attempt to respond back to the agent it received a transmission from, if at all. All responses are made concurrently. Again, for successful responses received above the SINR threshold, the data structure updating procedure occurs again.

\begin{algorithm}[t]
 \caption{ \normalsize Data Distribution Protocol }
  \label{alg:protocol}
\begin{algorithmic}
 \REQUIRE $\mathcal{A}, \pi, f, \Delta t $
  \WHILE{\textbf{true}} 
    \STATE \textbf{Transmission phase begins}, $t \Leftarrow t + \Delta t$ 
  \STATE Agents update current state in local data structures \\
  $\quad M_t^{i,i} = \bbx_t^i, T_t^{i,i} = t \quad \forall i \in \mathcal{A}, \hspace{1mm} t \geq 0$
  \STATE Agents evaluate transmission policy using local data structures\\
  $\quad p(k \in S_{t}^i) = \pi\left(\lbrace T_t^{i,j}, M^{i,j}_t, P^{i,j}_t,L^{i,j}_t  \rbrace_{j \in \mathcal{A}}  \right) \;\;\; \forall i, k \in \mathcal{A} $
  \STATE Transmissions are subject to interference \\
  $\quad \lbrace p(j \in R_t^i)  \rbrace_{i \in \mathcal{A}} = f\left(\lbrace S_{t}^i , \bbx_t^i \rbrace_{i \in \mathcal{A}} \right)$
  \STATE Agents update local data structures, $\forall i,k \in \mathcal{A}, j \in R_{t}^i $\\
  $\quad  T_t^{i,k} < T_t^{j,k} \Longrightarrow $
  \STATE $\quad\quad (T_t^{i,k} = T_t^{j,k})  \wedge  (M_t^{i,k} = M_t^{j,k})   \wedge (P_t^{i,k} = P_t^{j,k}) $
  \STATE \textbf{Response phase begins}, $t \Leftarrow t + \Delta t$
  \STATE Recipients transmit responses  \\
    $\quad  (j \in R_{t}^i )\land (j \in S_{t}^i)  \Longrightarrow i \in \bar{S}_{t}^j, \quad \forall j \in \mathcal{A} $
  \STATE Responses are subject to interference \\
    $\quad \lbrace p(i \in \bar{R}_t^j)  \rbrace_{j \in \mathcal{A}} = f\left(\lbrace \bar{S}_{t}^j , \bbx_t^j \rbrace_{j \in \mathcal{A}} \right)$
  \STATE Agents update local data structures, $\forall i,k \in \mathcal{A}, j \in \bar{R}_{t}^i $\\
  $\quad  T_t^{i,k} < T_t^{j,k}  \Longrightarrow $
  \STATE $\quad\quad (T_t^{i,k} = T_t^{j,k})  \wedge  (M_t^{i,k} = M_t^{j,k})   \wedge (P_t^{i,k} = P_t^{j,k}) $`
  \ENDWHILE
 \end{algorithmic} 
 \end{algorithm}

\subsection{Reinforcement Learning}



Finally, with the communication policy and protocol in place, we turn to our attention to finding a solution. To solve the problem in \eqref{eq:minimize_aoi} via reinforcement learning, we use the Proximal Policy Optimization algorithm \cite{schulman2017proximal} and parametrize both the policy and value functions as graph neural networks to take advantage of the modular nature of the data distribution task.


Our policy and value models are two separate neural networks, both comprised of a non-linear Aggregation GNN, with $f_{\text{enc}}$, $f_{\text{dec}}$ and $GN$ as 3 layer MLPs with 64 hidden units, and a ReLU activation only after the first two layers. 	

For the policy, $f_{\text{out}}$ is a linear function that reduces the high-dimensional latent space vectors to the required low-dimensional output, the logits of a Boltzmann distribution. Using the Gumbel-softmax, we use the logits to generate a discrete distribution over potential actions. At each timestep, each robot samples from this distribution to decide the recipient of its transmission. Also included as an action is the option for the agent not to transmit at the current timestep. For the value function, $f_{\text{out}}$ outputs 1 scalar per agent, which are summed to compute the team's value estimate.

Unless noted otherwise, we use a receptive field of 5 across all experiments. Each model was trained using an open source implementation of PPO \cite{hill2018stable} using $2 \times 10^{6}$ observations. We use an Adam optimizer with step size $1 \times 10^{-4}$ that is decayed by a factor of 0.95 for every 500 steps, and a batch size of 64. All input features were normalized by the mission area and the duration of the mission prior to being input to the neural network.


\section{Simulations} \label{sec:simulations}

To evaluate our data distribution system, we conducted simulations with a robot team performing three main tasks. The first two tasks serve to establish performance baselines and involve the agents passing information while stationary and while moving with random velocities. In the third task we use our system in a flocking maneuver, where each agent sets its own velocity according to its knowledge of the velocity of the rest of the team. In this case, global information about the velocity of each member of the team must be distributed in a timely manner so that the team converges on a common velocity before any agents can wander off. In each case, we report the age of information achieved by our system compared with a suite of other data distribution approaches outlined in Section \ref{sec:baselines}.

\subsection{Agent Dynamics}

In simulation we treat each robot as a point mass in $\mathbb{R}^2$ with first-order dynamics described by:
\begin{align}
\bbp_{t+1}^i = \bbp_{t}^i + \bbv_{t}^i \Delta t, \quad
\bbv_{t+1}^i = \bbv_{t}^i + \bbu_{t}^i \Delta t  
\end{align}
where for agent $i$ at time $t$, $\bbp_t^i \in \mathbb{R}^2$ is the agent's position  and $\bbv_t^i \in \mathbb{R}^2$ is the agent's velocity. An agent's local state is described by the 4-dimensional column vector $\bbx_t^i := \lbrack \bbp_t^i; \bbv_t^i \rbrack$ for agent $i$ at time $t$. This is the local state the agent transmit to others.

For the second task involving random motion, we use control inputs sampled uniformly according to $\bbu^i_t \sim \mathcal{N}(0, A_{max} / 3)$.	The control inputs for this task do not rely on the performance of the data distribution algorithm. In addition, agent velocities are clipped to the interval $(-V_{max}, V_{max})$ and velocities are also reflected when agents come into contact with the boundaries of the mission area, $(-P_{max}, P_{max})$, described in the following section.

\begin{figure}[t]
\centering
\includegraphics[width=1.0\textwidth]{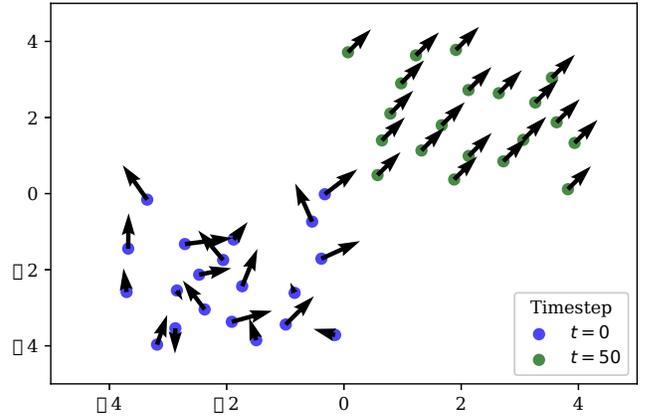}
\vspace{.11cm}
\vspace{-.32cm}
\caption{After starting with random velocities, agents communicate and eventually come to a global consensus. 
\vspace{-0.25cm}
}\label{fig:flock_vis}
\end{figure} 

In the flocking task, agents seek consensus in velocities, as can be seen in example in Fig. \ref{fig:flock_vis}. This task is sensitive to latency in the communication among fast-moving agents, so data distribution is essential for it success \cite{tolstaya2019learning}. Agent $i$ runs a controller based on its observations of the team after the initial timestep $T^{i,j}_0$, $\mathcal{F}_t^i = \lbrace j \in \mathcal{A} \text{ s.t. } T^{i,j} > T^{i,j}_0  \rbrace$, which computes the difference between the agent's current velocity and the average velocities among all observed agents:
\begin{align}
\bbu^i_t := 
\frac{1}{|\mathcal{F}^i_t|} \sum_{j \in \mathcal{F}^i_t} 
\begin{bmatrix}
0 & 0 & 1 & 0 \\
0 & 0 & 0 & 1\\
\end{bmatrix}
\times
\left( M^{i,j}_t - \bbx_t^i \right) 
\end{align}
%
This is a variant of the task proposed in \cite{tanner2003stable} but without control of agent spacing.

For a model trained to execute data distribution for flocking, we can quantify not only the age of information cost in \eqref{eq:minimize_aoi}, but we can also benchmark performance on the flocking task itself \cite{tolstaya2019learning}, quantifying the variance in agents' velocities: 
\begin{equation} \label{eq:flockcost}
\min_\pi \;\;  \frac{1}{|\mathcal{A}|} \sum_{t\in\mathcal{T}} \sum_{i \in \mathcal{A}} \left\| \bbv_t^i - \frac{1}{|\mathcal{A}|} \bigg[  \sum_{j \in \mathcal{A}} \bbv_t^j \bigg]  \right\|^2
\end{equation}
These two performance metrics for the flocking task are heavily correlated, but this may not be true for other tasks, in which low-latency information may be less important.

\subsection{Mission Specification}

The team of 40 robots operates in a mission area of 1 $\text{km}^2$, where robots' initial positions were generated uniformly at random within the mission space. For teams of other sizes, the mission area is changed to maintain a fixed agent density of 40 agents per 1 $\text{km}^2$.  For mobile agents, the velocity is provided as a ratio of the maximum distance the agent could travel across the mission area during the mission. The default velocity ratio is 0.15, corresponding to  $V_{max}=3$ m/s. The maximum acceleration is set to be $A_{max}=20 \text{m/s}^2$ for all experiments with mobile teams. 
    
We allot a communication time window of 100ms, in which both a request and response is made. In the case of random flooding, two requests are made. Each mission is 500 steps long, allowing 500 rounds of communication among agents.
The default maximum communication range for agents is 250 meters, which we plot in normalized units as 0.25 of the mission distance. Team positions were re-initialized until the communication graph had algebraic connectivity at the initial timestep, assuming a disk communication model with a range of 250 m. Across all experiments, we maintain a SINR threshold of 1 dBm, an additive white Gaussian noise of -50 dBm, and a system center frequency of 2.4 GHz.
In Section \ref{sec:results}, we benchmark the mean performance of all methods over 100 episodes, and provide the standard error of the mean using the error bars. 
    
\subsection{Baselines} \label{sec:baselines}

As mentioned in Section \ref{sec:intro}, data distribution is a regular part of many ad-hoc routing protocols. Thus, we compare our learned approach to three representative baselines inspired by such protocols: random flooding, round robin, and minimum-spanning tree.

 

Flooding is perhaps the most popular data distribution method used in ad-hoc routing protocols. The simplest form involves repeatedly rebroadcasting a message originating at a source node until it has reached every agent in the network. More advanced variants seek to improve inefficiencies by introducing rules governing which nodes should participate in rebroadcasting. In our simulations, we employ random flooding, which instructs agents to rebroadcast a message with some probability $p$ \cite{tseng2002broadcast}. While flooding in ad-hoc protocols often happens periodically to update network topology information, data distribution in the context of robot teams happens continuously, as the state information of each agent is constantly evolving. Thus, for our problem, the choice a node makes is whether to broadcast the latest information it has gathered, including what it has received from neighbors (i.e. rebroadcasting), or stay silent and allow other nodes to transmit. We allow random flooding to complete two transmissions over during our protocol's transmission-response window.

The second method we compare against is round robin. In this approach, a central agent is chosen as a base station and at each timestep,  one other agent exchanges data structures with it, reminiscent of a time-division multiple access approach \cite{miao2016fundamentals}. Unlike our approach and random flooding, round robin requires centralized coordination to identify the base station agent and schedule communications of other agents with the base station. Round robin follows the same transmit-response format used by our approach with $S_t^i$ always populated by the base station.

Finally, we also implement a Minimum Spanning Tree (MST) baseline where each agent attempts to exchange information with their parent in the spanning tree with probability $p$, remaining silent otherwise. This baseline seeks to capitalize off the fact that a MST minimizes the total edge length required to connect all the agents in the network. As such, the edges of the tree offer an efficient way to pass information throughout the network. One limitation of this approach is that it requires global knowledge of all agents' locations from the start of the episode. At a given timestep, the transmissions between child and parent follow the transmit-response format used by our method.



\section{Results}\label{sec:results}

First, we highlight the impact of the GNN’s
receptive field on its performance on the data distribution task, and characterize the impact of transmission power on the overall task difficulty. Next, we examine how the GNN can generalize to larger team sizes. Finally, we benchmark the GNN as the data distribution for a latency-sensitive flocking task.

\subsection{Stationary Teams}

\begin{figure}[t]
\centering
\includegraphics[width=0.9\textwidth]{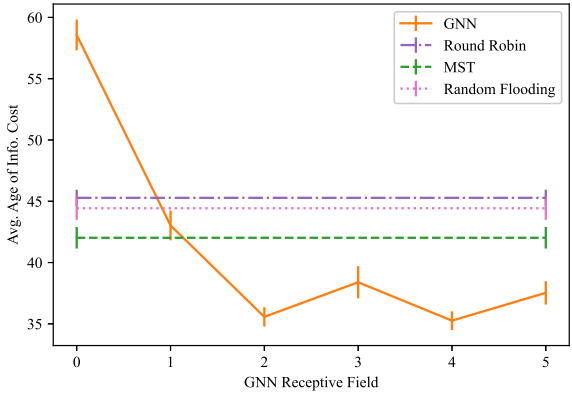}
\caption{For stationary agents, as the GNN receptive field is increased for the linear and non-linear models, AoI performance improves and exceeds MST and Round Robin. 
\vspace{-0.25cm}
}\label{fig:hops}
\end{figure}

Our first experiment investigates the effect of receptive field on the performance of the GNN controller as shown in Fig. \ref{fig:hops}.  We observe a general trend of improved performance with an increase in the receptive field of the architecture, which allows the model to reason about clusters of agents in its buffer. The GNNs with receptive fields of two or greater surpass the performance of Round Robin, MST and Random Flooding baselines.

\begin{figure}[t]
\centering
\includegraphics[width=0.9\textwidth]{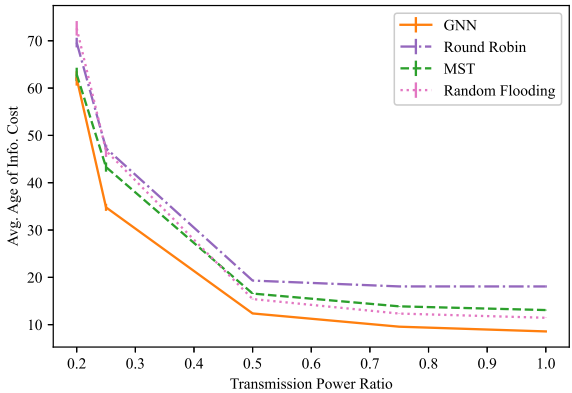}
\caption{For stationary agents, as agent transmission power increases, the data distribution task incurs lower latency on average for both the baseline controllers and the learner.
\vspace{-0.25cm}
}\label{fig:power}
\end{figure} 

Fig. \ref{fig:power} demonstrates the effect of varying the transmission power ratio, which is the fraction of the mission distance that an agent can successfully transmit across in the absence of interference. As the maximum transmission range increases, the problem becomes exponentially easier. However, many multi-robot teams operate in settings where hardware or environmental constraints restrict transmission distance. In these settings, especially when agents can only communicate at 25\% of the mission distance, our protocol performs best, having the greatest margin from the baseline approaches. The introduction of structure in how data propagated throughout the network allows our protocol to establish known routes of data flow. Instead of relying on direct communication with agents from the other side of the network, our protocol learns to leverage the established routes and often only needs short-range communications to access these.

\subsection{Mobile Teams}

\begin{figure}[t]
\centering
\includegraphics[width=0.9\textwidth]{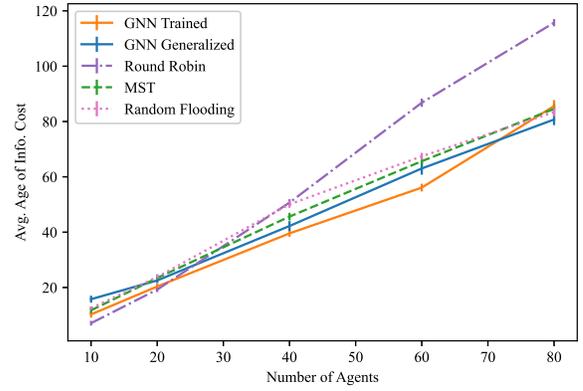}
\caption{GNN models trained on teams of 40 agents performing a random walk with a velocity ratio of 0.15 generalize well to larger team sizes.
\vspace{-0.5cm}
}\label{fig:mobile_n}
\end{figure}

Next, we examine the performance of the GNN in data distribution for mobile teams.
Fig. \ref{fig:mobile_n} demonstrates the generalization of a model trained on 40 mobile agents (GNN Generalization) to team sizes of 10 up to 80. For comparison, we also show the performance of a model trained on the corresponding team size (GNN Trained). We find that both models exceed the baselines for large team sizes. The transference property of Graph Neural Networks allows the model trained on a team of 40 to scale to teams of up to 80. With team sizes greater than or equal to 80 agents, we find that the $\mathcal{O}(|\mathcal{A}|^2)$ memory requirements make training difficult, and our generalized model begins to perform best.

\begin{figure*}[t]
\centering
\begin{subfigure}{0.495\columnwidth}
\centering
\includegraphics[width=0.9\textwidth]{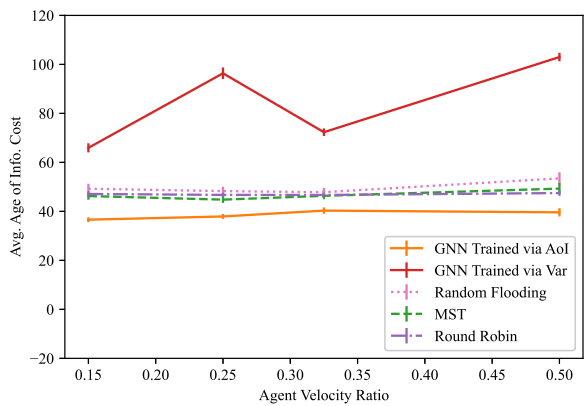}
\caption{Testing Age of Information Cost}
\label{fig:flocking_aoi}
\end{subfigure}
\hfill
\begin{subfigure}{0.495\columnwidth}
\centering
\includegraphics[width=0.9\textwidth]{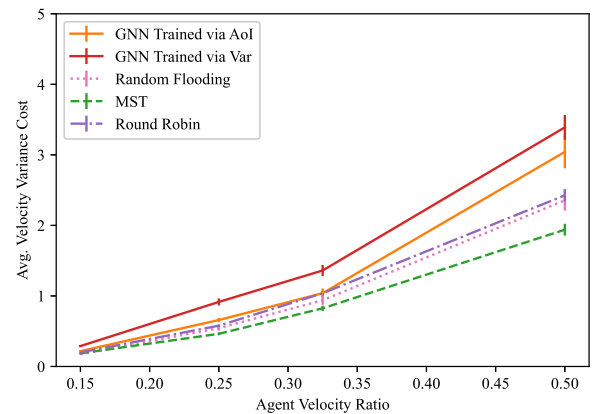}
\caption{Testing Velocity Variance Cost}
\label{fig:flocking_var}
\end{subfigure}
\caption{As agent velocity increases in the flocking task, the learner's performance begins to decline. Testing the velocity variance, the controller trained using the AoI reward performs better than the controller trained using Velocity Variance cost. \vspace{-1.0cm} 
} \label{fig:flocking}
\end{figure*}

Fig. \ref{fig:flocking} examines the flocking application and compares GNN models trained with the task-specific velocity variance cost \eqref{eq:flockcost} against models trained with the age of information cost \eqref{eq:minimize_aoi}.  Models trained with Age of Information cost perform better on the flocking task than models trained via the variance of velocities cost because the Age of Information provides a more direct signal for communication decisions. For using reinforcement learning for other problems with a data distribution component, age of information may be a highly-informative reward signal that could be used in conjunction with the task-specific reward function.
We observe that at higher velocities, the performance of both GNN controllers begins to stray away from the baselines because the initial route discovery phase is more challenging and agents may stray completely out of range before establishing communications. Providing information about the initial configuration of the team may be able to allow the GNN to improve performance from the initial time step. 

%



\section{Conclusions}\label{sec:conclusions}

 %

This work introduces a task-agnostic, decentralized method for data distribution in ad-hoc networks. We demonstrate that maintaining routes of data dissemination can help agents reason about multi-hop communication, resulting in more up-to-date network information as compared to industry-standard approaches. This performance extends to mobile robot teams, and transfers well for larger team sizes and the flocking task. We envision our protocol being applied to any collaborative multi-agent task reliant on sharing time-sensitive local observations across a wireless network.

In future work, we would like to run physical experiments with asynchronous execution. Another natural extension of this project is to not only consider when and with whom to communicate, but also what to communicate. At each time step, each agent would decide which portions of their data structures are essential to communicate for mission success, thus reducing payload size. 

\bibliographystyle{IEEEtran}
\bibliography{myIEEEabrv,references}
\end{document}